\documentclass[conference,a4paper]{APSIPA2021}
\usepackage{soul}
\usepackage[numbers]{natbib}
\usepackage[colorlinks,citecolor=black]{hyperref}
\usepackage{multirow}
\usepackage{amsmath}
\usepackage[]{amssymb}
\usepackage{amsfonts}
 \usepackage{booktabs}
\usepackage{amsxtra}
\usepackage{threeparttable}
\usepackage{graphicx}
\usepackage{float}
\usepackage{bm}
\usepackage{mlmath}
\usepackage{geometry}
\usepackage[normalem]{ulem}
\useunder{\uline}{\ul}{}
\usepackage{makecell}
\usepackage{tikz}
\usepackage{subfigure}
\usepackage{float}
\usepackage{lipsum}
\usepackage{makecell}

\usepackage{bbding}

\begin{document}
\title{Class Unbiasing for Generalization in Medical Diagnosis}
\author{
Lishi~Zuo, Man-Wai Mak, Lu~Yi, and Youzhi~Tu\\
\texttt{Dept. of Electrical and Electronic Engineering,}\\
\texttt{ Hong Kong Polytechnic University, Hong Kong SAR, China}\\
\texttt{E-mail: lishi.zuo@connect.polyu.hk, man.wai.mak@polyu.edu.hk,}\\
\texttt{lu-louisa.yi@connect.polyu.hk, 918tyz@gmail.com}
}

\maketitle
\thispagestyle{empty}
\begin{abstract}
Medical diagnosis might fail due to bias. In this work, we identified class-feature bias, which refers to models' potential reliance on features that are strongly correlated with only a subset of classes, leading to biased performance and poor generalization on other classes.
We aim to train a class-unbiased model (Cls-unbias) that mitigates both class imbalance and class-feature bias simultaneously. Specifically,  we propose a class-wise inequality loss which promotes equal contributions of classification loss from positive-class and negative-class samples.
We propose to optimize a class-wise group distributionally robust optimization objective---a class-weighted training objective that upweights underperforming classes---to enhance the effectiveness of the inequality loss under class imbalance.
Through synthetic and real-world datasets, we empirically demonstrate that class-feature bias can negatively impact model performance. Our proposed method effectively mitigates both class-feature bias and class imbalance, thereby improving the model's generalization ability.\looseness-1

\textbf{Keywords:} Class-feature bias, Class imbalance, Medical Diagnosis, Generalization
\end{abstract}

\section{Introduction}\label{sec:intro}

Medical data sets are often biased, which compromises the accuracy and reliability of models. A common bias is class imbalance, where models tend to prioritize the majority class at the expense of the minority classes, leading to poor performance on the latter. 
To mitigate this issue, researchers have proposed various methods, such as oversampling or undersampling classes~\cite{rahman2013addressing,depaudionet}, or using class weights to adjust the loss function~\cite{vSIDD,rahman2013addressing,RA-GCN}.

However, a critical oversight in current classification methods is the models' potential reliance on features that are strongly correlated with only a subset of classes, leading to biased performance and poor generalization on other classes. Even worse, if such learned correlations are spurious, i.e., correlated by chance rather than causality, the performance on the test set can be catastrophically poor.
For example, a model trained on a dataset comprising various biomedical measures of diabetic patients and healthy controls may mistakenly use the body mass index (BMI) to diagnose diabetes. While BMI is strongly correlated with the health conditions that cause diabetes, it is not a good indicator (biomarker) for diagnosing diabetes because people with high BMI may not necessarily have diabetes.
Specifically, diabetic patients often have a high BMI, i.e., $ P(H \mid D) > P(L \mid D) $, where $ H $ and $ L $ denote high and low BMI, respectively, and $ D $ indicates the presence of diabetes. 
However, this relationship between $P(H|D)$ and $P(L|D)$ is not necessarily true. Instead, it depends on a number of factors, including the population's health conditions and the demography of the subjects in the dataset.
As a result, all scenarios---$ P(H \mid D) > P(L \mid D) $, $ P(H \mid D) < P(L \mid D) $, or $ P(H \mid D) = P(L \mid D) $---are theoretically possible.
Therefore, a classifier that relies heavily on the BMI to make decisions has a high chance of classifying non-diabetic samples as diabetic.
This phenomenon can be particularly problematic when dealing with imbalanced datasets or complex feature spaces. 
In this paper, we define this issue as \textbf{class-feature bias} and focus on mitigating it in binary classification for medical diagnosis, where models tend to prioritize \textbf{class-specific features} that are informative for individual classes. Our goal is to develop \textbf{class-feature unbiased} models that leverage discriminative features shared across all classes (\textbf{class-shared features}).

Notably, the existence of class-feature bias is a major issue in medical diagnosis, where the primary procedure lies in identifying the presence or absence of specific biomarkers or their measured values that distinguish between disease and non-disease states. Leveraging class-shared features that are informative to both disease and non-disease states can help increase diagnosis accuracy and understand disease mechanisms. 
Conversely, neglecting these shared features risks overlooking critical indicators of disease progression or response to treatment, potentially leading to misdiagnosis, delayed treatment, or inadequate monitoring.
Therefore, developing models that can effectively learn and leverage class-shared features is essential for improving the accuracy and effectiveness of medical diagnoses.\looseness-1

Class-feature bias can be even more problematic under class imbalance scenarios, where one class dominates the dataset, further exacerbating the issue of biased performance.
In this study, we aim to train a \textbf{class-unbiased model (Cls-unbias)} that mitigates both class imbalance and class-feature bias simultaneously.
Leveraging insights from information theory, the key intuition is to ensure the features learned by the model are informative for both classes.
Specifically, we propose a class-wise inequality loss that equalizes the classification losses contributed by the samples from the positive and negative classes, respectively.
To improve the effectiveness of the class-wise inequality loss under class imbalance, we combine the cross-entropy loss with a class-wise group distributionally robust optimization (G-DRO) objective~\cite{g-dro}, which adaptively upweights the high-loss class.\looseness-1

We summarize our contributions as follows:

\begin{enumerate}
    \item To the authors' best knowledge, this study is the first to treat the class-feature bias as a research problem in medical diagnosis, empirically demonstrating its detrimental effect on binary classification performance.
    \item We provide both theoretical and empirical evidence showing the effectiveness of the class-wise inequality loss in addressing class-feature biases in balanced and imbalanced datasets.
    \item For practical considerations, we propose to use G-DRO to enhance the inequality loss's performance under class imbalance.
    \item We conducted extensive experiments on synthetic and real-world datasets. Our results demonstrate that promoting class-unbiased classification consistently improves model performance across diverse tasks and datasets with different input modalities.
\end{enumerate}

\section{Background \& Related Work}
\subsection{Empirical Risk Minimization}
Empirical risk minimization (ERM) has been widely adopted as a fundamental framework for training deep learning models. The goal is to minimize the average loss: 
\begin{align}
\mathcal{L}_\text{ERM} = \frac{1}{|\mathcal{B}|}\sum_{(\bm{x},y)\in\mathcal{B}}\ell(\bm{x},y),\label{eq:1-pure-erm}
\end{align}
where $\mathcal{B}$ denotes a batch of data and $\ell(\bm{x}, y)$ is the loss (e.g., cross-entropy loss for classification) for the input-target pair $(\bm{x}, y)$.
Variants of ERM, such as class-weighted ERM and per-class ERM, help mitigate the impact of the dominant classes under class-imbalance~\cite{erm-imb}.
In particular, class-weighted ERM modifies the loss function by applying class-dependent scaling factors $\alpha_y$'s to adjust for class imbalance:
\begin{align}
    \mathcal{L}_\text{ERM (cls-w)} = \frac{1}{|\mathcal{B}|}\sum_{(\bm{x},y)\in\mathcal{B}}\alpha_y\ell(\bm{x},y).
\end{align}
Per-class ERM computes the loss separately for each class and averages them:
\begin{align}
  \mathcal{L}_\text{ERM (per-cls)} = \sum_{c=1}^C \frac{1}{|\mathcal{B}^c|} \sum_{(\bm{x},y)\in\mathcal{B}^c} \ell(\bm{x}, y) = \sum_{c=1}^C\mathcal{L}^c,  
\end{align} 
where $\mathcal{B}^c$ denotes the samples in class $c$ in a training batch.
Per-class ERM ensures equal emphasis on all classes. Despite these enhancements, challenges remain in balancing performance across diverse distributions and mitigating overfitting in underrepresented classes.

\subsection{Bias in Medical Diagnosis}
Inherent biases in medical datasets are often invisible, causing models trained by ERM to produce inaccurate predictions.
Apart from the class imbalance mentioned in Section~\ref{sec:intro}, there are other
biases that arise from spurious features, including patient-specific
characteristics (e.g., age~\cite{hip}, gender~\cite{hip,gender_bias}, or speaker identity~\cite{vSIDD}) and environment-specific factors (e.g., equipment types~\cite{hip}, equipment manufacturers~\cite{hip}, or variations of diagnosis equipment across hospitals~\cite{pneumonia}).
To address these issues, researchers recommend using balanced and diverse training sets and reducing the influence of spurious features in model predictions~\cite{gender_bias,vSIDD,hip}. However, class-feature bias has largely been overlooked. This paper highlights the problem of class-feature bias in medical diagnosis and proposes a solution.\looseness-1

\subsection{Domain Generalization}
Domain generalization aims to improve model robustness under distributional shifts, where the distribution of unseen test data differs from that of the training data. 
Existing approaches to address the distributional shift mainly fall into two complementary categories: 
1) reduce domain discrepancies by making the data representations from different domains similar, using techniques such as data augmentation~\cite{DGaug-CrossGrad,DGaug-ada,DGaug-sdr}, feature alignment~\cite{dgalign-mtae,dgalign-er,dgalign-dlow}, or gradient matching~\cite{fish,fishr}; 
and 2) remove domain-specific information/factors, typically through disentanglement~\cite{DGdist-matchdg,DGdist-csd}, invariant risk minimization~\cite{irm,ib-irm,sparse-irm}, parameter decomposition~\cite{DGdist-undobias}, or sample reweighting~\cite{g-dro}.
For example, UndoBias~\cite{DGdist-undobias} enables support vector machines (SVMs) to handle dataset bias by learning both shared and dataset-specific weights, allowing the SVMs to separate dataset-specific biases from general object features. Invariant risk minimization (IRM)~\cite{irm,sparse-irm} aims to learn predictors that generalize across diverse environments by promoting invariance in the learned representations.

Our study aligns with the second category. Closely related prior work includes Risk Extrapolation (REx)~\cite{rex} and Group Distributionally Robust Optimization (G-DRO)~\cite{g-dro}. Specifically, REx~\cite{rex} minimizes the variance of training risks to achieve consistent performance across domains, encouraging loss equality at the domain level. 
In contrast, our work focuses on promoting loss equality at the class level to mitigate class-feature bias, and we establish its theoretical foundation using information-theoretic principles. G-DRO~\cite{g-dro}, which is designed for optimizing the worst-case group performance, is employed to address the challenges posed by extreme class imbalance.

\section{Class-feature Bias}
In this section, we formally define class-feature bias as models’ reliance on features that are correlated with only a subset of classes, leading to biased performance and poor generalization on other classes. Mathematically, let the dataset be $\mathcal{D} = \{ (\bm{x}_i, y_i) \}_{i=1}^N$, where $\bm{x}_i$ denotes the input features and $y_i \in \mathcal{C} = \{ k \}_{k=0}^{K-1}$ is the class label, with $\mathcal{C}$ the full set of class labels. Assume that some of the $K$ classes can be merged to form one class, resulting in $S$ jointed classes. For any subset of classes $\mathcal{C}^s \subseteq \mathcal{C}$, where $s\in\{1, \dots, S\}$, we define the corresponding data subset as $\mathcal{D}^{s} = \{ (\bm{x}, y) \in \mathcal{D} \mid y \in \mathcal{C}^s \}$. Then, we have $S$ disjoint subsets $\{\mathcal{D}^s\}_{s=1}^S$ such that $\mathcal{D} = \bigcup_{s=1}^{S} \mathcal{D}^s$, where $\bigcup$ denotes the union of subsets. 

Based on the Bayesian theory, if $Y$ and $X$ are independent, i.e., $P(X,Y)=P(X)P(Y)$, then we have $P(Y|X) = \frac{P(X,Y)}{P(X)} = P(Y)$, where the uppercase letters indicate random variables. Therefore, if there exists a subset $\mathcal{D}^{m}$ satisfying
\begin{align}
P(y\in \mathcal{C}^{m} | \bm{x})=P(y\in \mathcal{C}^{m}),\,\text{for some}\;m \in \{1,\dots,S\},\label{eq:cond1}
\end{align}
and at least one other subset $\mathcal{D}^{n}$ satisfying
\begin{align}
P(y\in \mathcal{C}^{n} | \bm{x})\neq P(y\in \mathcal{C}^{n}),\,n\in\{1,\dots S\}\setminus m,\label{eq:cond2}
\end{align}
with $P(\cdot)$ denoting the probability, then we say that $\bm{x}$ can cause class-feature bias. That is, $\bm{x}$ is informative for predicting samples from classes in $\mathcal{C}^{n}$, but it is irrelevant to classes in $\mathcal{C}^{m}$. Specifically, Eq.~\ref{eq:cond1} indicates that $\vx$ carries no information about whether a sample comes from $\mathcal{D}^{m}$; in other words, observing $\bm{x}$ does not increase the probability that the sample comes from $\mathcal{D}^{m}$.\looseness-1

Class-feature bias has received little attention in the literature. In~\cite{dics}, a special case in multi-class classification of class-feature bias is identified, where $\mathcal{C}^{n}$ consists of a single class and $\mathcal{C}^{m}$ includes a subset or all of the remaining classes. 
However, the authors did not provide a formal or general definition of class-feature bias, and their focus differed from ours. In particular, the authors aimed to keep features of the same class similar and features of different classes distinct by using a memory queue. However, this approach adds complexity and is less effective on small or low-diversity datasets. In contrast, our method is simple and well-suited for training binary classifiers for medical diagnosis using small medical datasets.

In this study, we first highlight class-feature bias as a significant research problem in medical diagnosis. We focus on binary cases where $\mathcal{C}^{m}=\{0\}$ and $\mathcal{C}^{n}=\{1\}$, or alternatively, $\mathcal{C}^{m}=\{1\}$ and $\mathcal{C}^{n}=\{0\}$. The Appendix provides a simple example that empirically demonstrates the class-feature bias in the binary case.
\section{Methodology}
\subsection{Class-wise Loss Equality}\label{sec:loss-sim}
Let $ \mathcal{D} = \{\mathcal{D}^{\text{pos}} \cup \mathcal{D}^{\text{neg}}\} $ be a dataset, divided into positive-class and negative-class samples. $ P^{\text{pos}}(Y | X) $ and $ P^{\text{neg}}(Y|X) $ are the true conditional distributions for samples from the positive and negative classes, respectively. 
$ \hat{P}(Y| X)$ is the predicted conditional distribution from a model, shared for both classes. 
Our objective is to ensure that the feature $ X $ provides a consistent reduction in uncertainty when predicting both positive and negative samples. This consistency indicates that $ X $ is informative for distinguishing the samples in the two classes, which can be expressed mathematically as:
\begin{align}\label{eq:entropy_eq}
    \mathbb{E}_{\mathcal{D}^{\text{pos}}}H(Y \mid X) = \mathbb{E}_{\mathcal{D}^{\text{neg}}}H(Y \mid X),
\end{align}
where $ \mathbb{E}_{\mathcal{D}^{\text{pos}}}H(Y|X) $ and $ \mathbb{E}_{\mathcal{D}^{\text{neg}}}H(Y|X) $ represent the expected conditional entropies for positive and negative samples, respectively.\looseness-1

We express the relationship between the conditional entropy and classification loss as follows:
\begin{align}
      \mathbb{E}_{\mathcal{D}^{c}}H(Y \mid X) = \mathcal{L}^{c} + D_{\text{KL}}(P^{c}(Y \mid X) \parallel \hat{P}^{c}(Y \mid X)),
\end{align}
where $D_{\text{KL}}(P\parallel \hat{P})$ represents the Kullback–Leibler (KL) divergence between distributions $P$ and $\hat{P}$, $c \in \{\text{pos}, \text{neg}\}$, and $\mathcal{L}^{\text{c}}$ is the cross-entropy loss for samples from $\mathcal{D}^{c}$, that is:
\begin{align}
   \mathcal{L}^{c}=-\mathbb{E}_{\mathcal{D}^c} P^{c}(Y | X) \log \hat{P}^{c}(Y | X). 
\end{align}
Ideally, if the model's predictions are perfect, i.e., $\hat{P}^{c}(Y | X) = P^{c}(Y | X) $, the KL divergence becomes zero, and in that case, when the condition in Eq.~\ref{eq:entropy_eq} is met and $P^{c}(Y | X)$ is well estimated, we will have
\begin{align}\label{eq:loss_eq}
    \mathcal{L}^{\text{pos}} = \mathcal{L}^{\text{neg}}.
\end{align}
In general, the condition in Eq.~\ref{eq:loss_eq} cannot be met, which is referred to as \textbf{class loss inequality}. This inequality can be caused by the following conditions:
\begin{enumerate}
    \item \textbf{Class-feature bias}. The model relies on features $X_{\text{spec}}$ that are specific to the class $c$. In that case, we will have $\mathcal{L}^c<\mathcal{L}^{\neg c}$, where $\neg$ denotes negation. As a result, the model can fail to identify samples from class $\neg c$ in the test set.
    \item \textbf{Class imbalance}. The model may under-utilize information from $X$ for the minority class $c$, leading to $\mathcal{L}^c>\mathcal{L}^{\neg c}$.
\end{enumerate}
To address the above problems, we propose to alleviate the inequality between the losses contributed by the positive- and negative-class samples by minimizing a class-wise inequality loss:
\begin{align}\label{eq:loss-sim}
    \mathcal{L}_\text{cls-ineq} = |\mathcal{L}^{\text{pos}} - \mathcal{L}^{\text{neg}}|.
\end{align}
The minimization of this loss leads to class-unbiased models.
The objective $ \mathcal{L}_\text{cls-ineq} $ is designed to minimize $ \mathcal{L}_\text{max} $ and maximize $ \mathcal{L}_\text{min} $, where $\mathcal{L}_\text{max} = \max(\mathcal{L}_\text{pos}, \mathcal{L}_\text{neg})$ and $\mathcal{L}_\text{min} = \min(\mathcal{L}_\text{pos},\mathcal{L}_\text{neg})$. Although maximizing $\mathcal{L}_\text{min}$ may seem counterintuitive, it serves an important role: it encourages the model to discard class-specific features that are highly informative for one class only (resulting in a low loss) but uninformative for the other class (leading to a high loss).

Intuitively, a large asymmetry or inequality between class-wise classification losses indicates that the model is overfitting to class-specific features, which can harm generalization. By reducing this inequality, the model is guided toward learning features that generalize across classes.

\subsection{Group Distributionally Robust Optimization for Class-imbalance}
In medical datasets, class imbalance often results in unreliable modeling of data distributions, which leads to difficulties in accurately estimating the posterior probability $ P^{c}(Y | X) $. 
As discussed in Section~\ref{sec:loss-sim}, class imbalance may cause the majority class to dominate in the training process, causing unequal contributions from the positive and negative classes. The unequal contributions will, in turn, lead to biased and degraded generalization performance in the resulting model.

Specifically, when the inequality between $\mathcal{L}^c$ and $\mathcal{L}^{\neg c}$ arises due to class imbalance, our objective should be to minimize $\mathcal{L}_\text{max}$, which will in turn minimize the loss of the majority class. However, $\mathcal{L}_\text{cls-ineq}$ also encourages maximizing $\mathcal{L}_\text{min}$, which can be counterproductive—particularly under severe class imbalance—by unintentionally penalizing the minority class and destabilizing the optimization.
This effect is particularly evident in the early training process, where the minority class receives insufficient attention. To alleviate the detrimental effect of class imbalance, we complement the cross-entropy loss with a G-DRO objective~\cite{g-dro}, which adaptively emphasizes the under-explored, underperforming class by assigning higher weight to the minority class, mitigating the imbalance and helping the $ \mathcal{L}_\text{cls-ineq} $ loss work effectively under class-imbalanced situation.
Specifically, the G-DRO objective is:
\begin{align}
    \mathcal{L}_\text{g-dro} = \sum_{c\in\{\text{pos},\text{neg}\}}\text{sg}\left(\frac{\exp({\tau\mathcal{L}^c})}{\exp({\tau \mathcal{L}^c})+\exp({\tau \mathcal{L}^{\neg c}})}\right)\mathcal{L}^c, 
\end{align}
where $\text{sg}(\cdot)$ denotes the stop-gradient operator, i.e., the enclosed term is treated as a constant during backpropagation.
Here, $\tau$ is a temperature hyper-parameter, which controls the extent to which the model focuses on the worst-performing class (the class with the largest loss).
Specifically, when $ \tau $ is large (e.g., 10), the model places greater emphasis on improving the performance of the worst class; conversely, when $ \tau $ is small (e.g., $10^{-2}$), the model treats both classes more equally.
Notably, when $ \mathcal{L}^{c} = \mathcal{L}^{\neg c} $, the group distributionally robust objective $ \mathcal{L}_\text{g-dro} $ simplifies to the per-class ERM.

While the class-wise inequality loss $ \mathcal{L}_\text{cls-ineq} $ can help mitigate the negative effects of class imbalance by minimizing $ \mathcal{L}_\text{max} $ to some extent, it becomes less effective when the model is already biased. Therefore, under scenarios where class imbalance significantly impacts training, we recommend using $ \mathcal{L}_\text{g-dro} $ to obtain a better initial estimate of $ \mathcal{L}_\text{pos} $ and $ \mathcal{L}_\text{neg} $, providing a stronger reference for optimizing $ \mathcal{L}_\text{cls-ineq} $.
Overall, the total loss is defined as:
\begin{align}\label{eq:cls-unbias}
    \mathcal{L}_\text{total} = \alpha \mathcal{L}_\text{cls-ineq} + \mathcal{L}_\text{g-dro}.
\end{align}
\section{Experimental Setup}
We evaluated the effectiveness of the class unbiased method on high-dimensional data (speech and image) in the real world. We provide a comprehensive analysis on the impact of class unbiasing on the class imbalance problem and the class-feature bias problem.
\subsection{Datasets}
The proposed method was evaluated on five medical datasets, including three speech datasets and two imaging datasets. Table~\ref{tab:dataset_stats} shows the detailed information about the number of samples in the positive and negative classes in these datasets.
\subsubsection{Speech Datasets}
\paragraph{DAIC-WOZ} 
We utilized the Distress Analysis Interview Corpus Wizard-of-Oz (DAIC-WOZ) dataset~\cite{daic_dataset}, a widely-used benchmark for depression diagnosis. The dataset comprises clinical interviews from 189 participants,
totaling approximately 30 hours of conversation data. Each interview was transcribed and annotated with the participants' depression status, as determined by the PHQ-8 scale~\cite{PHQ-8}. According to the official division of DAIC-WOZ dataset, we trained the models on the training set, validated on the development set, and tested on the test set. Pretrained wav2vec features~\cite{w2v} were extratced and preprocessed as stated in \cite{vSIDD}.

\paragraph{MODMA} 
The Multimodal Open Dataset for Mental Disorder Analysis (MODMA)~\cite{modma_dataset} is a depression dataset collected and released by Lanzhou University from the Second Hospital of Lanzhou University, containing audio recordings from 52 individuals.\footnote{\url{https://modma.lzu.edu.cn/data/index/}} 
In MODMA, the positive class constitutes 43\% of the data. We followed the splits of training, validation, and test set in \cite{mywork-cfaug}. The waveforms were segmented into 3.84-second segments, and 80-dimensional filterbank features (FBanks) were extracted as the model input.

\paragraph{ADReSS}
The ADReSS dataset~\cite{adress_dataset} is a speech-based dementia dataset, comprising speech recordings and their transcripts of participants describing the Cookie Theft picture from the Boston Diagnostic Aphasia Exam. The dataset includes 78 dementia participants and 78 healthy controls. Audio recordings were segmented to 3.84s segments, and 80-dimensional FBanks were extracted for training. 

\subsubsection{Imaging Datasets}
The MedMNIST benchmark~\cite{medmnistv2_dataset} provides medical imaging datasets designed for the rapid evaluation of machine learning methods on clinical tasks, including breast cancer diagnosis (BreastMNIST), diabetic retinopathy severity prediction (RetinaMNIST), colorectal cancer detection (PathMNIST), among others. We evaluated our method on the BreastMNIST and RetinaMNIST datasets, and we followed the official division of training, validation, and test sets.
\paragraph{BreastMNIST}
BreastMNIST contains 780 breast ultrasound images with a resolution of 28 × 28 pixels. It is designed for binary classification task where breast ultrasound images are categorized into ``positive" (normal and benign) and ``negative" (malignant). 

\paragraph{RetinaMNIST}
RetinaMNIST is designed for ordinal regression to predict the severity level of diabetic retinopathy, a diabetes-related eye disease, using retina fundus images. The labels correspond to five ordered levels of severity (0-4). We modified the dataset for binary classification by assigning levels 0 and 1 to the negative class, and levels 2-4 to the positive class.

\begin{table}[ht]
\centering
\caption{The number of positive and negative samples in the training and test sets of the five datasets used in this study.}
\label{tab:dataset_stats}
\setlength{\tabcolsep}{5.2mm}{
\begin{tabular}{c|cc|cc}
\hline
            & \multicolumn{2}{c|}{\textbf{Train}} & \multicolumn{2}{c}{\textbf{Test}} \\
            & \textbf{pos}     & \textbf{neg}     & \textbf{pos}    & \textbf{neg}    \\ \hline
DAIC-WOZ~\cite{daic_dataset}    &     31             &        76          &          14       &      33       \\
MODMA~\cite{modma_dataset}      &       9           &     12             &   13              &      14           \\
ADReSS~\cite{adress_dataset}    &     37             &      33            &           22      &       23          \\
BreastMNIST~\cite{medmnistv2_dataset} &         399         &       147           &        114         &      42           \\
RetinaMNIST~\cite{medmnistv2_dataset} &        260          &        820          &       88          &      312           \\ \hline
\end{tabular}}
\end{table}

\subsection{Model Structure}
\subsubsection{Speech Models}
The encoders of the speech models for the DAIC-WOZ, MODMA, and ADReSS datasets are three-layer fully connected (FC) networks with a $\texttt{tanh}$ activation function in their hidden layers. The network structures are [$\text{dim}^\text{in}$, $\text{dim}^\text{in} - \frac{\text{dim}^\text{in} - \text{dim}^\text{out}}{2}$, $\text{dim}^\text{out}$], where $\text{dim}^\text{in}$ is the dimension of the input features, and $\text{dim}^\text{out}$ denotes the output dimension (12, 16, and 16 for the DAIC-WOZ, MODMA, and ADReSS datasets, respectively). The frame-based outputs of the encoders were fed into a statistics pooling layer to obtain segment-level vectors by concatenating the vectors representing the mean across time, the standard deviation across time, and the mean first-order difference between successive feature frames. The pooled vectors were projected to one dimension via a linear layer to output the diagnosis state.

\subsubsection{Vision Models}
The encoders of the vision models follow the same experimental setting as described in \cite{mywork-cfaug}. The classifiers of the vision models are FC layers with dimensions [64, 128, 128, 1], using a $\texttt{ReLU}$ activation function in the hidden layers. 

\subsection{Evaluation}
We primarily focused on the F1-score and accuracy. As both the overall diagnostic accuracy and the model's performance on individual classes are important, especially in scenarios involving class-feature bias and class imbalance. We used the Macro-averaged F1-score (MF1), the F1-score for the positive class (pos-F1), and the F1-score for the negative class (neg-F1) as our evaluation metrics. Each experiment was repeated five times with different random initializations, and the reported results are the average performance across the five runs.

\subsection{Optimization}
The values of hyper-parameters are shown in Table~\ref{tab:hyper-param}.
The hyper-parameters $\alpha$ and $\tau$ were dynamically adjusted during training, changing linearly from $\alpha^\text{init}$ to $\alpha^\text{end}$, and from $\tau^\text{init}$ to $\tau^\text{end}$, respectively. Specifically, $\alpha$ was increased ($\alpha^\text{init} < \alpha^\text{end}$) during training to ensure that $\mathcal{L}_\text{cls-ineq}$ gradually took effect after $\hat{P}^{c}(Y | X)$ started to approximate $P^{c}(Y | X)$, followed by several rounds of model training. On the other hand, $\tau$ was decreased ($\tau^\text{init} < \tau^\text{end}$) to gradually shift the model's focus from exploring the worst class to treating the two classes equally. This adjustment of $\tau$ helps address extreme class imbalance early in the training, preventing failures where $\hat{P}^{c}(Y | X) \neq P^{c}(Y | X)$ due to class imbalance, which could degrade the effectiveness of $\mathcal{L}_\text{cls-ineq}$.

Cosine learning rate schedules were applied, where the learning rate $\eta$ was linearly increased from 0 during the initial $\omega$ iterations, then it was gradually decreased to 0 according to a cosine curve for the remaining iterations. 
The Adam optimizer was used for all experiments, with weight decay controlled by $\lambda$.
For experiments on the speech datasets, each iteration uses samples from one audio segment per unique speaker to ensure balanced speaker representation.\looseness-1

\begin{table}[ht]
\caption{Hyper-parameter settings for experiments. lr is learning rate, $B$ is batch size, $\omega$ is warm-up ratio, and $\lambda$ is the hyper-patameter controlling the regularization strength of weight decay.}
\label{tab:hyper-param}
\resizebox{\linewidth}{!}{
\begin{tabular}{c|ccccccc}
\hline
Dataset &
  lr &
  Iterations &
  $B$ &
  $\omega$ &
  $\lambda$ & $\alpha^\text{init}\to\alpha^\text{end}$
   & $\tau^\text{init}\to\tau^\text{end}$
   \\ \hline
DAIC-WOZ~\cite{daic_dataset}    & $3\times 10^{-3}$ & 100   & -   & 0.1 & 0    & 0$\to$1& 2$\to$0.01 \\
MODMA~\cite{modma_dataset}       & $1\times 10^{-3}$ & 200  & -   & 0.5 & 0    &  0$\to$1& 2$\to$0.01 \\
ADReSS~\cite{adress_dataset}      & $1\times 10^{-3}$ & 100   & -   & 0.1 & 0.01 & 0$\to$1 & 10$\to$1 \\
BreastMNIST~\cite{medmnistv2_dataset} & $3\times 10^{-4}$ & 200 & 128 & 0.5 & 0.01 &  0$\to$2& 10$\to$1 \\
RetinaMNIST~\cite{medmnistv2_dataset} & $1\times 10^{-4}$ & 150 & 64  & 0.1 & 0    &  1$\to$2& 10$\to$1 \\ \hline
\end{tabular}}
\end{table}

\section{Results}
\subsection{Main Results} 
The main results across five datasets are presented in Table~\ref{tab:main-res}. Overall, the proposed class unbiased model consistently outperforms the baselines (ERM \& ERM (cls-w)), across all datasets.

The models trained on most medical datasets suffer from class imbalance, with the issue being particularly severe on DAIC-WOZ where ERM even fails to learn meaningful patterns. While the common class-weighting strategy (ERM (cls-w)) alleviates this problem to some extent, its effectiveness remains limited. In contrast, the class unbiased model outperforms ERM (cls-w), demonstrating its superiority as a solution for class-imbalanced datasets. This improvement stems not only from better handling of class imbalance but also from mitigating potential class-feature bias.

Notably, on the ADReSS dataset, where the class distribution is nearly balanced, applying class weighting (ERM (cls-w)) even degrades performance. Meanwhile, the class unbiased model remains effective, suggesting the presence of unseen class-feature bias. 
This validates the importance of explicitly addressing class-feature bias and underscores the class-unbiased model's effectiveness in enhancing generalization and robustness, even in class-balanced scenarios.

In addition, we observe that the performance improvements of the class-unbiased model over the baselines are more pronounced on the speech datasets (DAIC-WOZ, MODMA, and ADReSS) than the image datasets (BreastMNIST and RetinaMNIST). This is likely because speech tasks are inherently more complex, making it harder for the models to capture true causal relationships, rendering them more susceptible to spurious correlations, including class-feature biases. As a result, the proposed class-unbiased model achieves greater improvements on speech datasets by effectively addressing these potential biases.

\subsection{Ablation Study}
This section investigates the effect of different losses on the DAIC-WOZ dataset, with the results shown in Table \ref{tab:ablation}. In the experiments, the model trained by ERM fails to learn meaningful patterns due to the severe class imbalance in the DAIC-WOZ dataset.
Specifically, the model collapses into always predicting the majority (negative) class, thereby effectively minimizing the cross-entropy loss and achieving around 70\% accuracy on the training set. Apparently, this seemingly decent performance is an illusion, as the model fails to detect the minority (positive) class. While the variants of ERM partially alleviate this issue—either through a class-weighting strategy (ERM (cls-w)) or by placing equal emphasis on all classes (ERM (per-cls))—the overall performance remains poor, indicating that the model is still negatively affected by class imbalance and potential class-feature bias. 

Solely using $\mathcal{L}_\text{g-dro}$ (Row 3) is a better approach than using the variants of ERM (ERM (per-cls) \& ERM (cls-w)) to handle the class-imbalance problem in the DAIC-WOZ dataset, as the former promotes the exploration of under-represented classes to reduce class loss inequality.
However, it fails to handle class-feature bias, because its strategy is to maximize exposure to the minority class rather than minimize reliance on biased class-specific features.
In contrast, the class-unbiased models (Rows 4 \& 5) substantially improve performance by reducing class loss inequality, thereby mitigating both class imbalance and class-feature bias through explicitly promoting class-wise loss equality, which leads to more generalizable and reliable performance.

Comparing Row 4 and Row 5 of Table~\ref{tab:ablation} further reveals additional improvement brought by incorporating $\mathcal{L}_\text{g-dro}$, demonstrating its effectiveness in helping the class-wise inequality loss achieve its intended effect. This is particularly evident on the DAIC-WOZ dataset, where severe class imbalance skews the model’s estimation of $P(Y|X)$, thereby limiting the effectiveness of $\mathcal{L}_\text{cls-ineq}$. By introducing $\mathcal{L}_\text{g-dro}$, we mitigate the distortion in the estimated $P(Y|X)$ caused by class imbalance, improving the overall performance.

\begin{table}[ht]
\centering
\caption{Main results on the five real-world datasets. MF1 (with $\pm 1$ standard deviation) for all experiments are reported.}
\label{tab:main-res}
\resizebox{\linewidth}{!}{
\begin{tabular}{c|ccccc}
\hline
 & \textbf{DAIC-WOZ} & \textbf{MODMA} & \textbf{ADReSS} & \textbf{BreastMNIST} & \textbf{RetinaMNIST} \\ \hline
ERM          & - & 0.714$\pm$0.039 & 0.493$\pm$0.035 & 0.814$\pm$0.020 & 0.647$\pm$0.035 \\
ERM(cls-w)   & 0.483$\pm$0.048 & 0.746$\pm$0.027 & 0.489$\pm$0.031 & 0.809$\pm$0.023 & 0.667$\pm$0.032 \\
Cls-unbias & 0.627$\pm$0.044 & 0.782$\pm$0.070 & 0.530$\pm$0.012 & 0.833$\pm$0.029 & 0.682$\pm$0.038 \\ \hline
\end{tabular}}
\end{table}

\begin{table*}[ht]
\caption{Ablation study on DAIC-WOZ.}
\label{tab:ablation}
\resizebox{\linewidth}{!}{
\begin{tabular}{c|c|cccc|cccc}
\hline
\multirow{2}{*}{\textbf{Row}}  
&\multirow{2}{*}{\textbf{Method}}                        
& \multicolumn{4}{c|}{\textbf{Valid}}              
& \multicolumn{4}{c}{\textbf{Test}} \\
&
&
\textbf{pos-F1} & \textbf{neg-F1} & \textbf{MF1} & \textbf{Acc}
& \textbf{pos-F1} & \textbf{neg-F1} & \textbf{MF1} & \textbf{Acc} \\ \hline
1&ERM (per-cls) & 0.522$\pm$0.057 & 0.769$\pm$0.037 & 0.646$\pm$0.033 & 0.691$\pm$0.033 &
0.228$\pm$0.025 & 0.642$\pm$0.066 & 0.435$\pm$0.023 & 0.515$\pm$0.056\\
2&ERM (cls-w) & 0.503$\pm$0.063 & 0.804$\pm$0.026 & 0.654$\pm$0.040 & 0.720$\pm$0.033 &
0.244$\pm$0.033 & 0.721$\pm$0.064 & 0.483$\pm$0.048 & 0.596$\pm$0.072 \\
3&$\mathcal{L}_\text{g-dro}$ & 0.670$\pm$0.048 & 0.844$\pm$0.018 & 0.757$\pm$0.030 & 0.789$\pm$0.023 &
0.318$\pm$0.038 & 0.701$\pm$0.04 & 0.510$\pm$0.020 & 0.587$\pm$0.037\\
4&ERM (per-cls) + $\mathcal{L}_\text{cls-ineq}$ & 0.693$\pm$0.035 & 0.845$\pm$0.008 & 0.769$\pm$0.019 & 0.794$\pm$0.011 & 
0.452$\pm$0.048 & 0.723$\pm$0.027 & 0.588$\pm$0.025 & 0.634$\pm$0.025  \\
5&$\mathcal{L}_\text{g-dro} + \mathcal{L}_\text{cls-ineq}$ &   0.717$\pm$0.043 &  0.858$\pm$0.025 &   0.788$\pm$0.031   & 0.811$\pm$0.029 &    
0.510$\pm$0.054 &   0.744$\pm$0.036   & 0.627$\pm$0.044  & 0.664$\pm$0.043     \\ \hline
\end{tabular}}
\end{table*}

\subsection{Case Study on Balanced and Imbalanced Datasets.}
\begin{figure*}[!th]
\centering
\subfigure[$\mathcal{L_\text{cls-ineq}}$\label{fig:cls-ineq-b}]{
\begin{minipage}[b]{0.235\textwidth}
\includegraphics[width=1\linewidth]{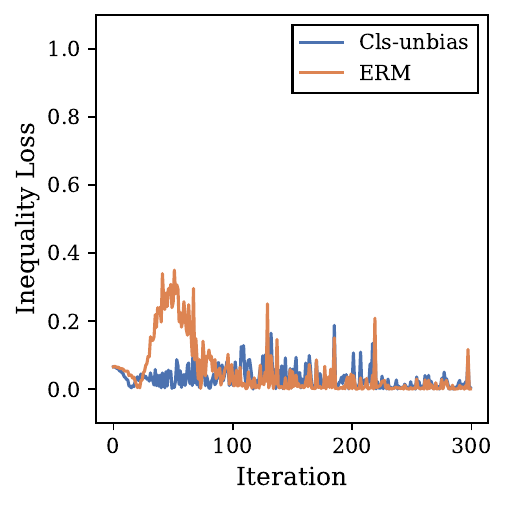}
\end{minipage}}
\subfigure[$\mathcal{L}^\text{pos}$ and $\mathcal{L}^\text{neg}$.\label{fig:nll-compare-b}]{
\begin{minipage}[b]{0.235\textwidth}
\includegraphics[width=1\linewidth]{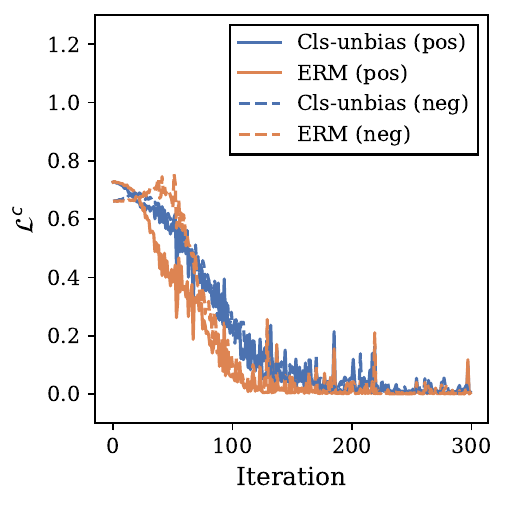}
\end{minipage}}
\subfigure[Classification loss $\mathcal{L}_\text{ERM}$ on validation set.\label{fig:valid-loss-b}]{
\begin{minipage}[b]{0.235\textwidth}
\includegraphics[width=1\linewidth]{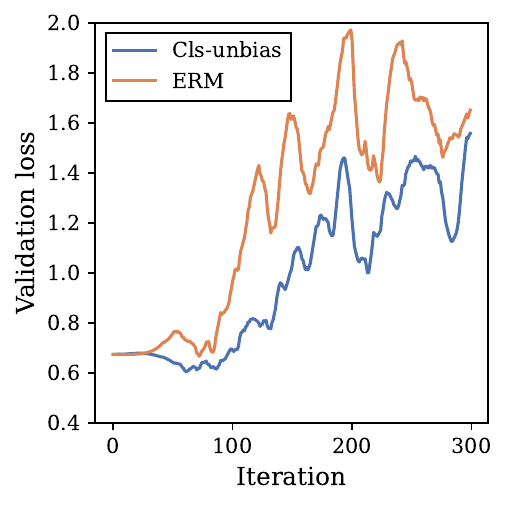}
\end{minipage}}
\subfigure[MF1 on validation set.\label{fig:case-mf1-b}]{
\begin{minipage}[b]{0.235\textwidth}
\includegraphics[width=1\linewidth]{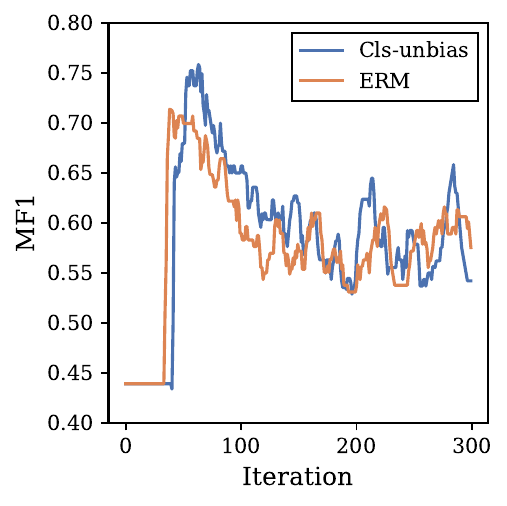}
\end{minipage}}
\caption{Case study on resampled balanced RetinaMNIST dataset. The number of training samples is 100.}
\label{fig:case-balanced}
\end{figure*}
\begin{figure*}[!th]
\centering
\subfigure[$\mathcal{L_\text{cls-ineq}}$\label{fig:cls-ineq-imb}]{
\begin{minipage}[b]{0.235\textwidth}
\includegraphics[width=1\linewidth]{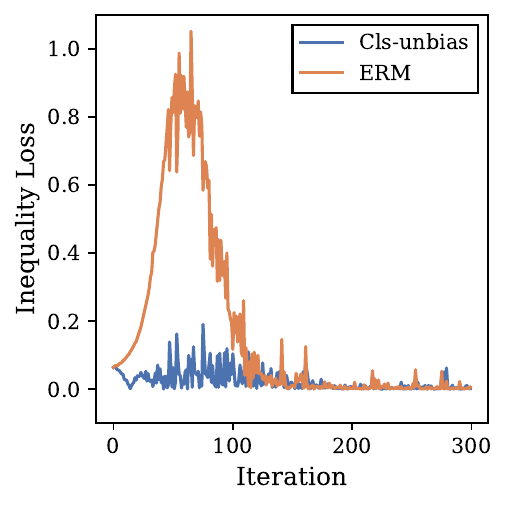}
\end{minipage}}
\subfigure[$\mathcal{L}^\text{pos}$ and $\mathcal{L}^\text{neg}$.\label{fig:nll-compare-imb}]{
\begin{minipage}[b]{0.235\textwidth}
\includegraphics[width=1\linewidth]{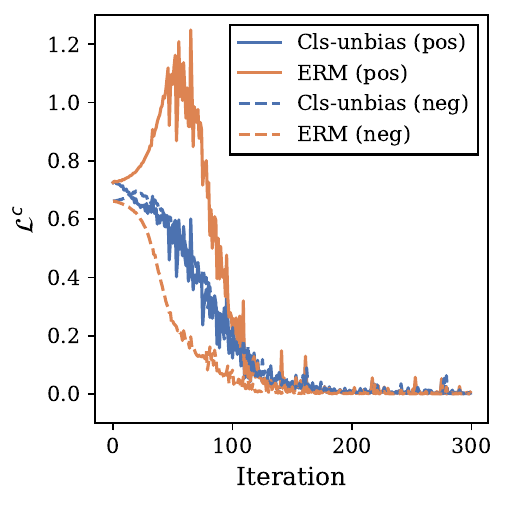}
\end{minipage}}
\subfigure[Classification loss $\mathcal{L}_\text{ERM}$ on validation set.\label{fig:valid-loss-imb}]{
\begin{minipage}[b]{0.235\textwidth}
\includegraphics[width=1\linewidth]{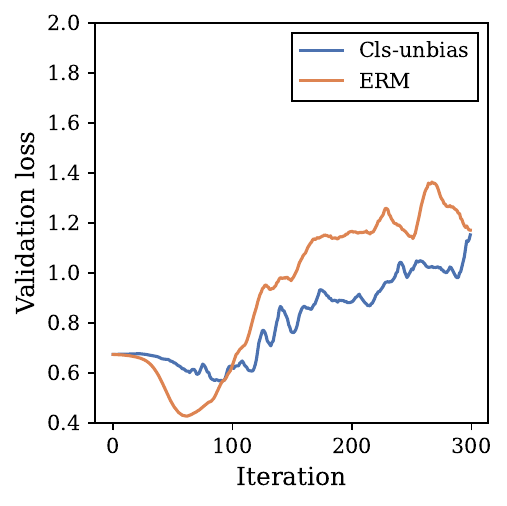}
\end{minipage}}
\subfigure[MF1 on validation set.\label{fig:case-mf1-imb}]{
\begin{minipage}[b]{0.235\textwidth}
\includegraphics[width=1\linewidth]{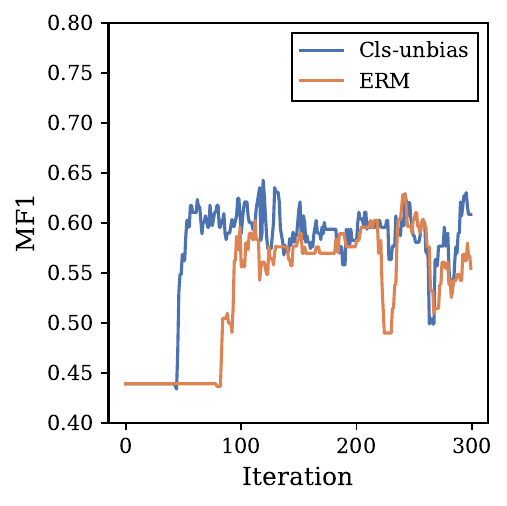}
\end{minipage}}
\caption{Case study on resampled imbalanced RetinaMNIST dataset. The number of training samples is 100. 25\% of training samples are from the positive class.}
\label{fig:case-imbalanced}
\end{figure*}

This section empirically investigates the effect of the proposed class-unbiased approach on balanced and imbalanced datasets. Specifically, we resampled the RetinaMNIST dataset to construct two settings: a balanced dataset with 100 samples (50\% positive) and an imbalanced dataset with 100 samples (25\% positive). The experimental results are shown in Table~\ref{tab:case-study}. The training processes on both balanced and imbalanced datasets are shown in Figures~\ref{fig:case-balanced} and~\ref{fig:case-imbalanced}, respectively.

We first note that the performance improvement of Cls-unbias on the balanced dataset is relatively modest, which is expected since only class-feature bias is addressed in a class-balanced setting.
In fact, we found that the degree of inequality varies across different network initializations—some initializations result in substantial inequality.
Figure~\ref{fig:case-balanced} illustrates an initialization that exhibits a significant class loss inequality in the balanced dataset, indicating the presence of class-feature bias.
As shown in the figure, reducing this inequality using Cls-unbias leads to lower validation loss (Figure~\ref{fig:valid-loss-b}) and higher validation MF1 scores (Figure~\ref{fig:case-mf1-b}). Moreover, the slower increase in validation loss suggests improved generalization and robustness.

In contrast, the effect of Cls-unbias on the imbalanced dataset is much more pronounced (Figure~\ref{fig:case-imbalanced}), underscoring its effectiveness in addressing both class imbalance and potential class feature bias. Notably, under class imbalance, the severity of class loss inequality is evident, as reflected by the higher values of $\mathcal{L}_\text{cls-ineq}$ and a larger gap between $\mathcal{L}_\text{pos}$ and $\mathcal{L}_\text{neg}$ compared to the balanced setting under ERM. Specifically, when the model overly focuses on the majority (negative) class, its loss decreases rapidly, while the minority (positive) class loss increases. This results in a lower overall validation loss but comes at the cost of under-representing the minority class and amplifying class loss inequality. Such inequality arises not only from class imbalance but also from class feature bias. 
Importantly, class feature bias can be exacerbated under class imbalance, as training on imbalanced datasets may encourage the model explore class-specific features, making the role of Cls-unbias even more critical.
Overall, the significant improvements observed on the imbalanced dataset highlight the approach’s effectiveness in mitigating both sources of bias.

\begin{table}[ht]
\centering
\caption{Experimental results on the resampled balanced and imbalanced RetinaMNIST datasets.}
\label{tab:case-study}
\begin{tabular}{c|c|c|cc}
\hline
\textbf{Row} & \textbf{Methods} & \textbf{cls-balanced} & \textbf{MF1} & \textbf{Acc} \\ \hline
1            & ERM              &   \Checkmark  &  0.621$\pm$0.030 &    0.719$\pm$0.041  \\
2            & Cls-unbias     &   \Checkmark  &     0.633$\pm$0.023  &  0.738$\pm$0.034 \\ \hline
3            & ERM              &  \XSolidBrush  &  0.568$\pm$0.036  &  0.732$\pm$0.026   \\
4            & Cls-unbias     &  \XSolidBrush  &   0.604$\pm$0.039  &   0.743$\pm$0.025   \\ \hline
\end{tabular}
\end{table}
\section{Discussions \& Conclusions}
In this paper, we identify class-feature bias and highlight it as a significant research problem in medical diagnosis. We proposed a class-wise inequality loss to learn a class-unbiased model that handles class-feature bias and class imbalance simultaneously. 
For practical considerations, 
we propose to optimize the model using class-wise G-DRO~\cite{g-dro} to enhance the effectiveness of inequality loss under class imbalance.
Experimental results shows that the proposed Cls-unbias approach consistently enhances the generalization ability of models under balanced and imbalanced scenarios, across diverse tasks and datasets with different input modalities.

\textbf{\textit{From binary classification to multi-class classification.}}
Extending Cls-unbias from binary classification to multi-class classification is feasible. Similar to \cite{rex}, one possible strategy is to reduce the variance of the class-wise losses. Encouraging class-wise loss equality remains effective when the number of classes is small. Determining the number of classes for which the multi-class extension of Cls-unbias remains valid can be left to future work.\looseness-1

However, as the number of classes increases, the effectiveness of Cls-unbias becomes limited. This is because it is increasingly difficult to identify features that are informative across all classes, as many classes may share many common attributes. Focusing on extracting such class-shared features may result in the model learning few or even no information for prediction, ultimately degrading performance. In such cases, alternative strategies such as matching approaches---for example, comparing extracted feature vectors with learned class prototypes---may be more appropriate. Nevertheless, the proposed definition of class-feature bias provides a useful theoretical foundation for guiding the design of multi-class classification methods, even in settings with a large number of classes.\looseness-1

Finally, we emphasize that Cls-unbias remains highly useful and important in medical diagnosis settings, i.e., binary or few-class classification, where strict feature selection and extraction are critical, as classification failures can lead to severe consequences, including misdiagnosis, delayed treatment, or inadequate monitoring. Moreover, misdiagnosis caused by class-feature bias may result in potential ethical harms. For instance, as discussed in the Section~\ref{sec:intro}, using BMI as a dominant predictive feature for diabetes may lead the model to incorrectly associate high BMI with the disease, reinforcing harmful stereotypes and potentially marginalizing individuals with atypical clinical presentations. Such biased representations can exacerbate healthcare disparities and erode trust in AI-assisted medical decisions.

\bibliographystyle{unsrt}

\section{Appendix}
\subsection{Example of Class-feature Bias}\label{sec:linear-eg}

Similar to \cite{fish}, we designed simple experiments using linear classifiers to demonstrate class-feature bias in binary classification under both class-balanced and class-imbalanced conditions. We showed that both class imbalance and class-feature bias can negatively impact classification performance and that making the classifier class-unbiased can improve generalization to out-of-distribution scenarios. The codes are available at \url{https://github.com/zuo-ls/cls-unbias}. 

\begin{figure*}[ht]
\centering
\subfigure[ $\mathcal{D}^{\text{tr}}$\label{fig:feat-dist-tr}]{
\begin{minipage}[b]{0.3\textwidth}
\includegraphics[width=1\linewidth]{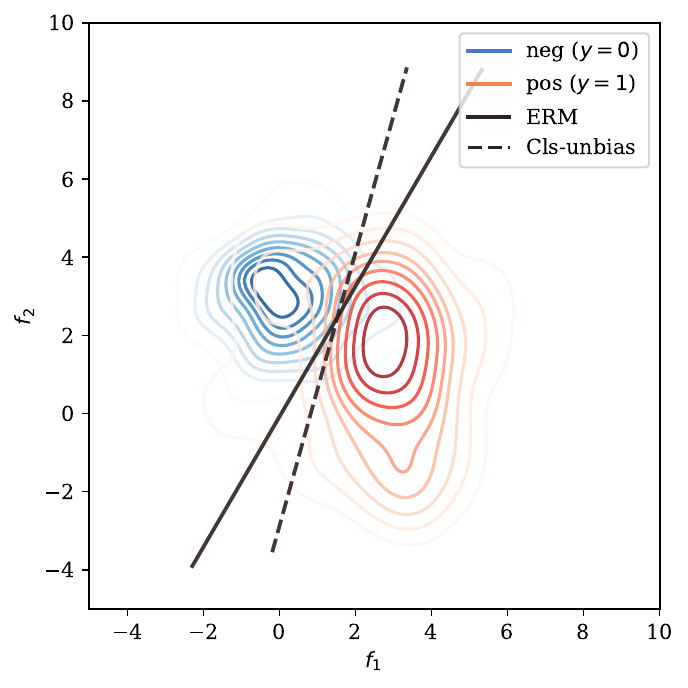}
\end{minipage}}
\subfigure[ $\mathcal{D}^{\text{test1}}$\label{fig:feat-dist-test}]{
\begin{minipage}[b]{0.3\textwidth}
\includegraphics[width=1\linewidth]{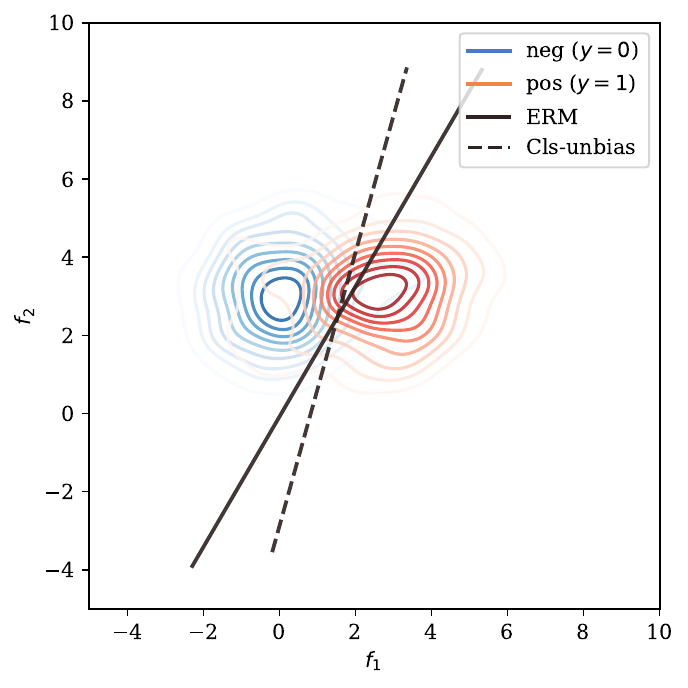}
\end{minipage}}
\subfigure[  $\mathcal{D}^{\text{test2}}$\label{fig:feat-dist-test2}]{
\begin{minipage}[b]{0.3\textwidth}
\includegraphics[width=1\linewidth]{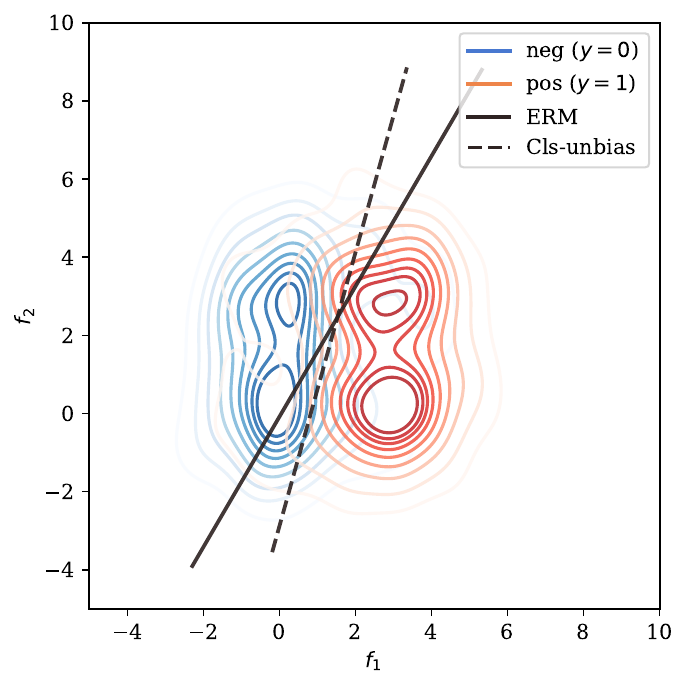}
\end{minipage}}
\caption{The decision boundaries of a linear classifier trained by the ERM (Eq.~\ref{eq:1-pure-erm}) and the proposed class-unbiased method (Eq.~\ref{eq:cls-unbias}) under the class-balance scenario. The contourplots show distributions of the two features ($f_1$ and $f_2$) for the positive (pos) and negative class (neg) for the three datasets: $\mathcal{D}^{\text{tr}}$, $\mathcal{D}^{\text{test1}}$, and $\mathcal{D}^{\text{test2}}$. The positive and negative classes have equal prior in $\mathcal{D}^{\text{tr}}$.}
\label{fig:feat-dist}
\end{figure*}

\begin{figure*}[ht]
\centering
\subfigure[  $\mathcal{D}^{\text{tr}}$\label{fig:feat-dist-tr-imb}]{
\begin{minipage}[b]{0.3\textwidth}
\includegraphics[width=1\linewidth]{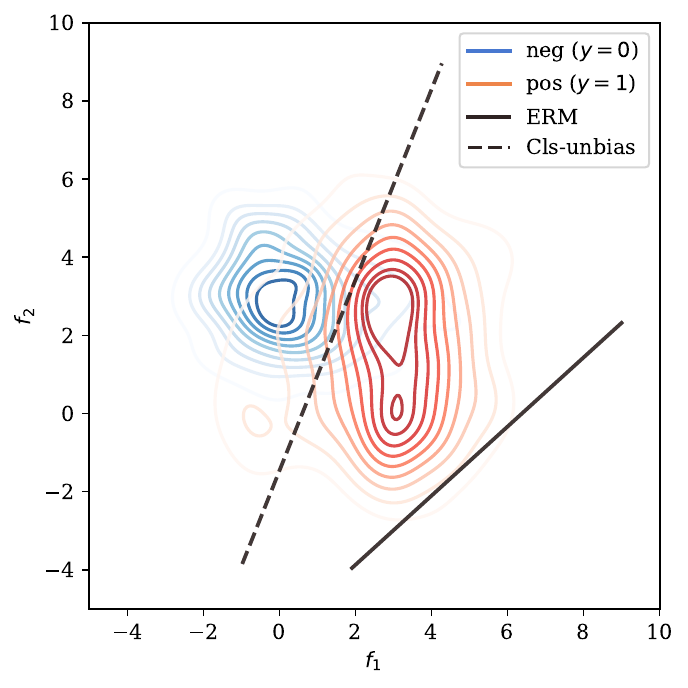}
\end{minipage}}
\subfigure[  $\mathcal{D}^{\text{test1}}$\label{fig:feat-dist-test-imb}]{
\begin{minipage}[b]{0.3\textwidth}
\includegraphics[width=1\linewidth]{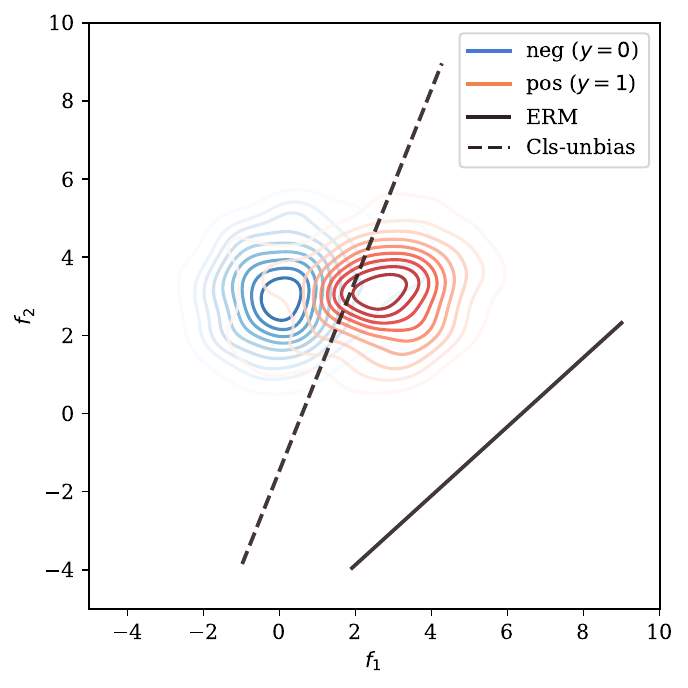}
\end{minipage}}
\subfigure[  $\mathcal{D}^{\text{test2}}$\label{fig:feat-dist-test2-imb}]{
\begin{minipage}[b]{0.3\textwidth}
\includegraphics[width=1\linewidth]{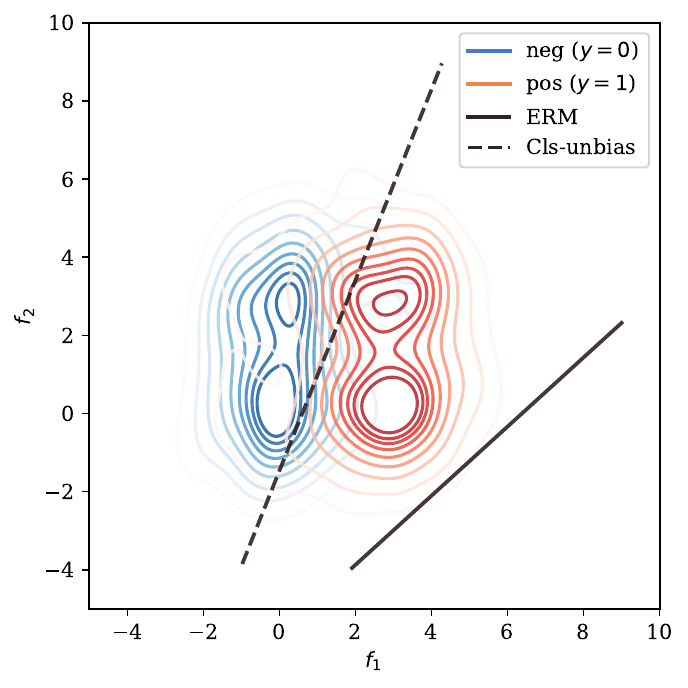}
\end{minipage}}
\caption{The decision boundaries of a linear classifier trained by the ERM (Eq.~\ref{eq:1-pure-erm}) and the proposed class-unbiased method (Eq.~\ref{eq:cls-unbias}) under the class-imbalance scenario. The contourplots show distributions of the two features ($f_1$ and $f_2$) for the positive (pos) and negative class (neg) for the three datasets: $\mathcal{D}^{\text{tr}}$, $\mathcal{D}^{\text{test1}}$, and $\mathcal{D}^{\text{test2}}$. The prior probability of the positive class is 0.1 in $\mathcal{D}^{\text{tr}}$, i.e., $P(y=1)=0.1$.}
\label{fig:feat-dist-imb}
\end{figure*}

Consider a 2D space containing feature vectors $\bm{x}$'s, where $\bm{x} = [f_1, f_2]^\top$ is an input feature vector containing two features $f_1$ and $f_2$, and $y \in \{0, 1\}$ is the classification label of $\bm{x}$. We set $f_1\in\{a,b\}$ and $f_2\in\{a,b\}$, where $a\sim \mathcal{N}(0,1)$ and $b\sim \mathcal{N}(5,1)$. Therefore, we may express $\bm{x} = [(a|b),(a|b)]^\top$, where $(a|b)$ means selecting either $a$ or $b$ from the set $\{a,b\}$. 
We applied this principle to generate three datasets, namely $D^\text{tr}$, $D^\text{test1}$, and $D^\text{test2}$. We used $D^\text{tr}$ to train a linear classifier and tested it on $D^\text{test1}$ and $D^\text{test2}$.
Fig.~\ref{fig:feat-dist} and Fig.~\ref{fig:feat-dist-imb} show the probability distributions of features $f_1$ and $f_2$ for the positive class ($y=1$) and negative class ($y=0$) under class-balanced and class-imbalanced scenarios, respectively.\looseness-1

\textbf{\textit{Class-specific Feature.}} Fig.~\ref{fig:feat-dist-tr} shows that $f_2$ is a class-specific feature in $\mathcal{D}^\text{tr}$. More specifically, for a classifier that uses $f_2$ as the only feature, an unknown sample having $f_2 \ll 5$ (say $f_2=-2$) indicates that it is likely a sample from the positive class (red). Meanwhile, assigning samples with $f_2\approx 5$ to the negative class (blue) can ensure that all negative samples are correctly classified. Therefore, $f_2$ is specific to the negative class.
On the other hand, Fig.~\ref{fig:feat-dist-test} and Fig.~\ref{fig:feat-dist-test2} show that $f_2$ is an irrelevant feature in $\mathcal{D}^\text{test1}$ and $\mathcal{D}^\text{test2}$, as its value does not determine the class of any unknown samples in these datasets.

\textbf{\textit{Spurious Feature.}} Applying the model trained in $\mathcal{D}^\text{tr}$ to the data in $\mathcal{D}^\text{test2}$ reveals that $f_2$ is a spurious feature. This is because although $f_2$ can partly determine the class labels in the samples in $\mathcal{D}^\text{tr}$, it fails to do the same in $\mathcal{D}^\text{test2}$.

\textbf{\textit{Class-shared Feature.}} In Fig.~\ref{fig:feat-dist}, $f_1$ is equally effective in classifying $\bm{x}$'s into the positive or negative classes for all datasets ($\mathcal{D}^\text{tr}$, $\mathcal{D}^\text{test1}$, and $\mathcal{D}^\text{test2}$), meaning that it is invariant to the differences in three datasets.
Also, $f_1$ is an example of class-shared features mentioned in Section~\ref{sec:intro}, because it can differentiate both the positive and negative classes.

In the ideal class-unbiased condition, the classofier should learn a vertical decision boundary in Fig.~\ref{fig:feat-dist} and Fig.~\ref{fig:feat-dist-imb} to separate the two classes. This means that the classifier pays full attention to $ f_1 $, while assigning no attention to $ f_2 $, resulting in high accuracy on both $\mathcal{D}^\text{test1}$ and $\mathcal{D}^\text{test2}$.
However, the actual decision boundaries of the linear classifier trained by the ERM (Eq. 1) and the proposed class-unbiased method (Eq.~\ref{eq:cls-unbias}) on balanced and imbalanced datasets are shown in Fig.~\ref{fig:feat-dist} and Fig.~\ref{fig:feat-dist-imb}, respectively. This discrepancy between the ideal and actual decision boundaries suggests the existence of class-specific features and spurious features.

Table~\ref{tab:res-toy} shows the classification performance of the linear classifier on the three datasets shown in Fig.~\ref{fig:feat-dist} and Fig.~\ref{fig:feat-dist-imb} under class-balanced and class-imbalanced scenarios, respectively. The classifier comprises two input nodes and one output node, with the normalized weights of the two input features indicated in the last column.\footnote{We omit the bias term of the classifier as it does not affect our discussion.}
To simulate the class-imbalanced and class-balanced situations, we set $p(y=1)=0.1$ in the class-imbalanced experiments (Rows 3 and 4) and $p(y=1)=0.5$ in the class-balanced experiments (Rows 1 and 2) in the training set. Meanwhile, we kept $p(y=1)=p(y=0)=0.5$ in the test sets for all experiments.

The key conclusions from the toy experiment are summarized below:
\begin{enumerate}
    \item The classifier trained by ERM (Row 1) relies heavily on the class-specific feature ($f_2$), as evidenced by the relatively large weights assigned to it. This reliance leads to degraded performance on $\mathcal{D}^\text{test1}$ for the positive class.
    The performance degradation becomes more severe under the class-imbalanced condition, as evident in Row~3.
    Worse, as mentioned earlier, because $f_2$ is a spurious feature for $\mathcal{D}^\text{test2}$, its performance on the biased class (negative class) can be significantly compromised~(Row~1).
    
    \item By comparing the weights in Row 3 with those in Row~1, we observe that class imbalance increases the model’s reliance on the unintended feature $ f_2 $. This leads to improved detection of the negative samples but significantly degrades the performance on the positive samples. 
    The model in Row~3 exhibits the combined effects of class-specific bias and class imbalance, which jointly distort the learned decision boundaries and amplify the asymmetric performance between the two classes.
    \item In all situations, Cls-unbias, which promotes the reduction of class loss inequality, can effectively alleviate the reliance on the class-specific feature $f_2$, while simultaneously improving performance. Moreover, it helps mitigate the class imbalance problem.
\end{enumerate}

\begin{table*}[ht]
\caption{Results for the experiments on the synthetic example. For the imbalanced dataset, $P(y=1)=0.1$. For meaningful comparisons of the attention to the features $f_1$ and $f_2$ by different models, we normalized the model's weights $\bm{w}=[w_1,w_2]$ by $\bm{w}/\|\bm{w}\|$ and show them in the ``Normalized Weights'' column.}
\label{tab:res-toy}
\setlength{\tabcolsep}{3.3mm}{
\begin{tabular}{c|c|c|cc|cc|cc|c}
\hline
\multirow{2}{*}{\textbf{Row}} &
  \multirow{2}{*}{\textbf{Method}} &
  \multirow{2}{*}{\textbf{Class-balanced}} &
  \multicolumn{2}{c|}{\textbf{Train}} &
  \multicolumn{2}{c|}{\textbf{Test 1}} &
  \multicolumn{2}{c|}{\textbf{Test 2}} &
  \multirow{2}{*}{\textbf{Normalized Weights}} \\
 &
   &
   &
  \textbf{neg-acc} &
  \textbf{pos-acc} &
  \textbf{neg-acc} &
  \textbf{pos-acc} &
  \textbf{neg-acc} &
  \textbf{pos-acc} &
   \\ \hline
1 &
  ERM &
  \Checkmark &
   0.848&
   0.850&
   0.880&
   0.703&
   0.679&
   0.825&
  [0.858, $\minus$0.514] \\
2 &
  Cls-unbias &
  \Checkmark &
   0.853&
   0.864&
   0.876&
   0.780&
   0.802&
   0.845&
  [0.962, $\minus$0.275] \\
3 &
  ERM &
  \XSolidBrush &
   1.000&
   0.001&
   1.000&
   0.000&
   1.000&
   0.004&
  [0.663, $\minus$0.749] \\
4 &
  Cls-unbias &
  \XSolidBrush &
   0.852&
   0.850&
   0.880&
   0.708&
   0.770&
   0.841&

  [0.925, $\minus$0.379] \\ \hline
\end{tabular}}
\end{table*}
\end{document}